\documentclass[sigconf, screen]{acmart}

\AtBeginDocument{%
  }
\setcopyright{none}      % 完全移除版权声明
\acmDOI{}               % 清空 DOI
\acmISBN{}              % 清空 ISBN
\acmConference{}        % 清空会议标题
\acmYear{}              % 清空会议年份
\copyrightyear{}        % 清空版权年份
\usepackage{fancyhdr}
\makeatletter
\def\@acmArticle@string{} % 清除 "Article" 标记
\def\@article@string{}    % 清除逗号残留
\makeatother

\usepackage{epsfig}
\usepackage{graphicx}
\usepackage{amsmath}
\usepackage{textcomp}
\usepackage{color}
\usepackage{subcaption}
\usepackage{booktabs}
\usepackage{tabularx}
\usepackage{multirow}
\usepackage{float}  %设置图片浮动位置的宏包
\usepackage{caption}
\usepackage{float} 
\usepackage{lipsum}
\usepackage{overpic} % 图片内引用
\usepackage{threeparttable}
\ifodd 1
\newcommand{\com}[1]{\textbf{\color{red}(COMMENT: #1)}} %comment of the 
\else

\newcommand{\com}[1]{}
\fi

\begin{document}

\title{EDmamba: Rethinking Efficient Event Denoising with\\ Spatiotemporal Decoupled SSMs}

\author{
Ciyu Ruan\textsuperscript{1,*}, 
Zihang Gong\textsuperscript{2,*}, 
Ruishan Guo\textsuperscript{1}, 
Jingao Xu\textsuperscript{3},
Xinlei Chen\textsuperscript{1, \dag}  \\
\textsuperscript{1}Tsinghua University,
\textsuperscript{2}Harbin Institute of Technology,
\textsuperscript{3}Carnegie Mellon University, \\
{\{softword77, gongzihang0201, ruishanguo314, xujingao13\}@gmail.com}, { chen.xinlei@sz.tsinghua.edu.cn}
}

% \affiliation{%

% \affiliation{%

% \affiliation{%

% \affiliation{%

% \affiliation{%

\renewcommand{\shortauthors}{Ruan and Gong, et al.}


\begin{abstract}
Event cameras provide micro-second latency and broad dynamic range, yet their raw streams are marred by spatial artifacts (e.g., hot pixels) and temporally inconsistent background activity. Existing methods jointly process the entire 4D event volume (x, y, p, t), forcing heavy spatio-temporal attention that inflates parameters, FLOPs, and latency. We introduce EDmamba, a compact event-denoising framework that embraces the key insight that spatial and temporal noise arise from different physical mechanisms and can therefore be suppressed independently. A polarity- and geometry-aware encoder first extracts coarse cues, which are then routed to two lightweight state-space branches: a Spatial-SSM that learns location-conditioned filters to silence persistent artifacts, and a Temporal-SSM that models causal signal dynamics to eliminate bursty background events. This decoupled design distills the network to only 88.9K parameters and 2.27GFLOPs, enabling real-time throughput of 100K events in 68ms on a single GPU, 36$\times$ faster than recent Transformer baselines. Despite its economy, EDmamba establishes new state-of-the-art accuracy on four public benchmarks, outscoring the strongest prior model by 2.1 percentage points.

\end{abstract}

\maketitle
\begingroup
\renewcommand\thefootnote{}\footnotetext{* Equal contribution. \\ \dag\ Corresponding author.}
\addtocounter{footnote}{0}
\endgroup
\section{Introduction}
\label{sec:intro}
Inspired by biological vision, event cameras asynchronously record per-pixel brightness changes with microsecond latency, ultra-high dynamic range ($>$120 dB), and low power consumption ($<$10 mW). These unique properties enable event-based perception to excel in high-speed and high-dynamic-range scenarios, powering breakthroughs in visual tracking~\cite{jiang2020object}, SLAM~\cite{gehrig2024low}, and obstacle avoidance~\cite{falanga2020dynamic}. However, this high temporal resolution is a double-edged sword: it also amplifies sensitivity to minor brightness fluctuations, thermal noise, and sensor imperfections. As a result, event streams often contain a large number of spurious events that obscure valid motion patterns, hinder downstream perception, and overwhelm system bandwidth with excessive event rates~\cite{gallego2020event}. Robust denoising is therefore a critical prerequisite for building scalable, high-performance event-driven systems.

\begin{figure}[t]
  \centering
  \includegraphics[width=\linewidth]{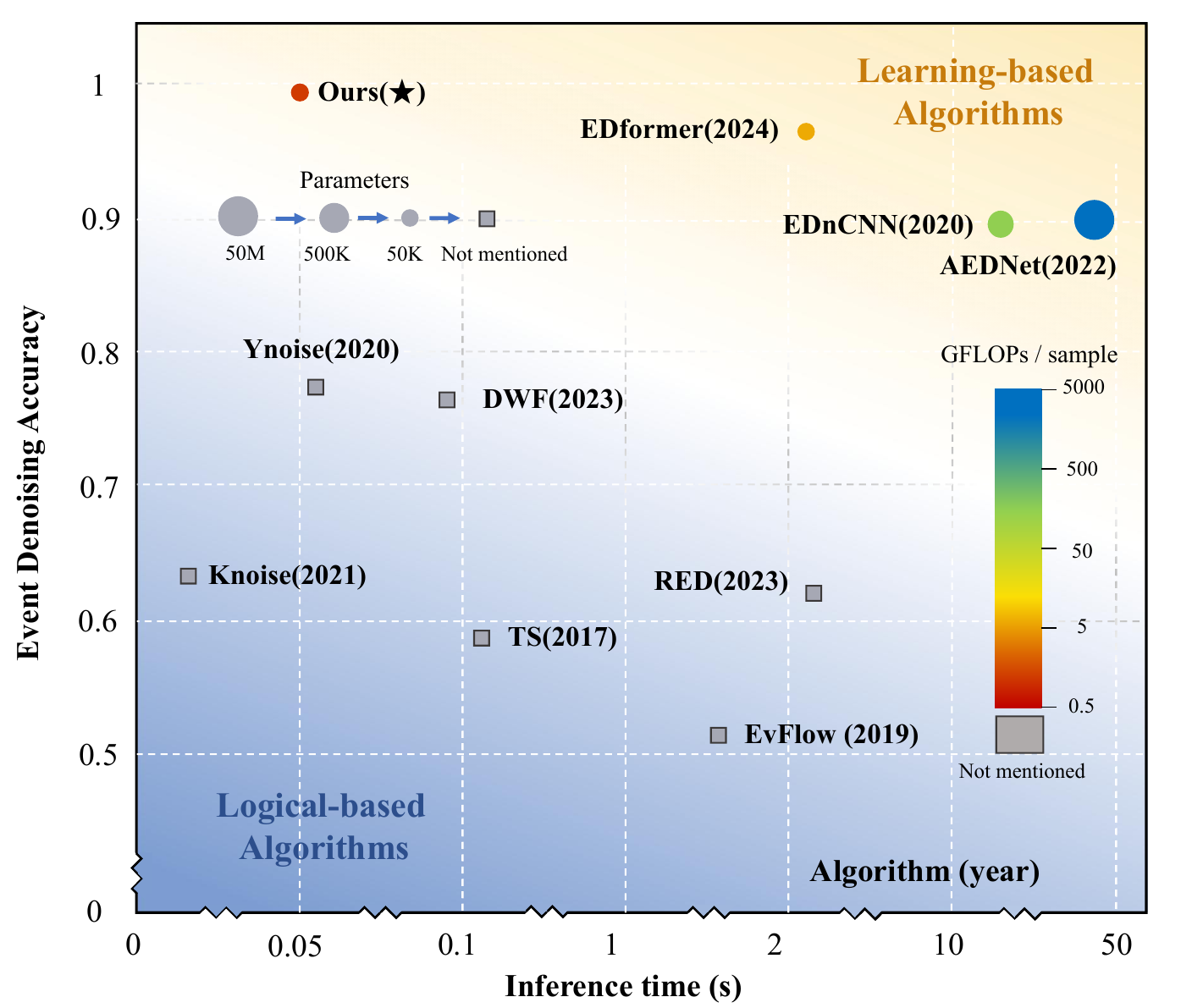}
  % \fbox{\rule{0pt}{2.5in} \rule{0.9\linewidth}{0pt}}
  \vspace{-0.7cm}
  \caption{Performance vs. efficiency on event denoising with 100K events from DND21 (346$\times$260, Hotel-bar and Driving scenes). EDmamba (red star) achieves state-of-the-art accuracy with low FLOPs, few parameters, and fast inference. All methods were evaluated under identical settings. Marker colors indicate FLOPs; sizes reflect parameter counts.}
  \vspace{-0.7cm}
  \label{fig:compasison}
\end{figure}

% Recent advances in event denoising have evolved from early statistical priors~\cite{feng2020event}, spatiotemporal filtering~\cite{baldwin2019inceptive}, and surface fitting~\cite{wang2019ev} to deep learning-based methods. The CNN-based method EDnCNN~\cite{baldwin2020event} and the point-based model AEDNet~\cite{fang2022aednet} focus on local patterns, whereas the Transformer-based method EDformer~\cite{jiang2024edformer} enables global modeling but suffers from quadratic complexity. More recently, state space models such as Mamba~\cite{gu2023mamba} offer linear-time sequence modeling with low memory overhead, and Pre-Mamba~\cite{ruan2025pre} extends this paradigm to 4D event deraining. Despite their architectural differences, these models typically rely on a single, heavy backbone for joint spatial-temporal modeling, introducing redundant computation and limiting adaptability to the heterogeneous nature of noise. Their high parameter counts and slow inference make them unsuitable for real-time applications; in practice, processing a 100K-event batch can take more than 2 seconds on Transformer-based models~\cite{jiang2024edformer}, which undermines the core advantage of event cameras in high-speed scenarios such as UAV navigation or autonomous driving.
Recent advances in event denoising have progressed from early methods based on statistical priors~\cite{feng2020event}, spatiotemporal filtering~\cite{baldwin2019inceptive}, and surface fitting~\cite{wang2019ev} to data-driven approaches. Deep learning-based models such as the CNN-based EDnCNN~\cite{baldwin2020event} and the point-based AEDNet~\cite{fang2022aednet} focus on local neighborhood patterns, while the Transformer-based EDformer~\cite{jiang2024edformer} enables global context modeling but suffers from intensive computation. More recently, state space models like Mamba~\cite{gu2023mamba} have emerged for linear-time sequence modeling, and Pre-Mamba~\cite{ruan2025pre} extends this framework to 4D event deraining.

Despite architectural diversity, these models commonly treat event denoising as a 4D problem across spatial and temporal dimensions, and adopt unified spatio-temporal processing backbones. This joint modeling necessitates dense  self- or cross-attention across space and time, resulting in redundant computation and limited adaptability to heterogeneous noise patterns. As a result, these models are often over-parameterized and suffer from high latency, limiting their suitability for real-time use. For instance, Transformer-based models can take over 2 seconds to process 100K events~\cite{jiang2024edformer}, which severely undermines the speed advantage of event cameras in high-throughput applications such as UAV navigation and autonomous driving.

To address this, we rethink event denoising from a noise-centric perspective. While event noise originates from diverse sources such as shot noise, fixed-pattern artifacts, and thermal leakage, its manifestations are often decoupled across time and space. Temporal noise, such as stochastic firings and polarity flips, lacks motion continuity and coherence, whereas spatial noise from hot pixels produces localized, structurally aberrant activations. This observation motivates a decoupled architecture that separates spatial and temporal denoising into two lightweight, parallel branches. By isolating distinct noise patterns, this design reduces computational overhead and improves adaptability, offering lower latency without sacrificing accuracy.

% To address this, we rethink event denoising from a noise-centric perspective. Although event noise arises from various sources such as shot noise, fixed-pattern noise, and thermal leakage, we observe that its manifestations are often independent across time and space. Temporally inconsistent noise, such as stochastic firings or polarity flips, lacks motion continuity, while spatially inconsistent noise caused by hot pixels leads to localized, structurally aberrant activations. This insight motivates a decoupled architecture, where spatial and temporal denoising are handled by two lightweight branches. 
% % Such decomposition simplifies each sub-task, supports parallel execution, and avoids heavy joint modeling, resulting in reduced latency without compromising denoising quality.
% Such decomposition simplifies each sub-task and enables parallel execution, reducing computational complexity compared to joint spatiotemporal modeling. As spatial and temporal noise typically arise from distinct physical mechanisms, this decoupled strategy maintains denoising fidelity while achieving lower latency. 

Building on this insight, we propose \textbf{EDmamba}, a lightweight and effective \textbf{E}vent \textbf{D}enoising model that combines modality-specific processing with the efficiency of State Space Models. EDmamba operates directly on raw event streams represented as compact 4D event clouds, which encode spatial coordinates, polarity, and precise timestamps. A Coarse Feature Extraction (CFE) module first encodes geometric and polarity-aware features. These are then processed by two separate SSM branches: a Spatial Mamba (S-SSM) that captures local geometric patterns to suppress spatially incoherent noise, and a Temporal Mamba (T-SSM) that models temporal dynamics to eliminate temporally inconsistent activations. While structurally decoupled, the two streams interact via a shared Spatial-Temporal State Space Block (STSSB), enabling joint reasoning without entangled feature extraction. A U-Net-style encoder-decoder backbone ensures multi-scale information flow through skip connections, supporting both localized denoising and global context integration.

EDmamba achieves state-of-the-art performance with a compact design. It requires only 88.98K parameters and 2.27 GFLOPs, and efficiently processes 100K events in just 0.0685 seconds. Compared to recent Transformer-based methods, it achieves a 2.08\% improvement in denoising accuracy while offering 36 times faster inference.  This balance of accuracy, speed, and scalability makes EDmamba a practical and robust solution for real-time event-based perception. Our main contributions are as follows:
\begin{itemize}
    \item We demonstrate that spatial and temporal noise in event streams exhibit distinct patterns and can be more effectively suppressed through decoupled denoising. This design insight enables simultaneous improvements in both accuracy and efficiency.
    % \item We propose EDmamba, the first state space model tailored for event denoising. The architecture includes a 4D Coarse Feature Extraction module and two lightweight, decoupled Mamba branches for spatial and temporal denoising, supported by a Spatial-Temporal State Space Block for joint reasoning.
    \item We propose EDmamba, the first state space model specifically designed for event denoising. It extracts polarity- and geometry-aware features from 4D event clouds, and applies two lightweight, decoupled Mamba branches that independently model spatial and temporal noise characteristics for targeted suppression.
    \item We conduct extensive experiments demonstrating that EDmamba outperforms strong baselines in both denoising accuracy and inference speed, while requiring significantly fewer parameters.
\end{itemize}

\section{Related Work}
\label{sec:related}
Event denoising has advanced through various approaches, including signal processing, statistical modeling, surface fitting, and deep learning. These methods have improved the robustness of event-based perception in noisy conditions.

\noindent \textbf{Statistical methods}.
Early techniques leveraged statistical heuristics to suppress spurious events by evaluating event density within local spatiotemporal neighborhoods, rejecting low-density events as noise~\cite{feng2020event}. The pioneering work by Delbrück~\cite{delbruck2008frame} proposed density-based filtering with spatial context. Subsequent improvements~\cite{liu2015design, khodamoradi2018n, guo2022low} optimized computational efficiency through enhanced event storage and processing. However, their reliance on manual parameter tuning hinders generalization across diverse scenarios.

\noindent \textbf{Filtering-based methods}.
To better accommodate the sparse, asynchronous nature of event data, researchers have introduced filtering techniques along temporal, spatial, and spatiotemporal axes. Temporal filters~\cite{baldwin2019inceptive} leverage temporal correlation, often exploiting patterns from edge motion. Spatial filters~\cite{ieng2014asynchronous} analyze local intensity changes to distinguish signal events. Spatiotemporal filters~\cite{czech2016evaluating, wu2020denoising} integrate both domains to suppress background activity (BA) noise while preserving motion-related information. Notably, \cite{liu2015design} showed that combining spatial and temporal cues yields more effective noise suppression than either alone.

\noindent \textbf{Surface fitting techniques}.
Surface fitting offers an alternative denoising strategy by modeling the spatiotemporal distribution of events. EV-Gait~\cite{wang2019ev} and GEF~\cite{duan2021guided} apply local plane fitting and optical flow to differentiate noise from coherent motion. Time-surface (TS) representations~\cite{lagorce2016hots, baldwin2019inceptive} convert event streams into decaying memory surfaces that encode temporal history, aiding in distinguishing structured signals from random outliers. These methods perform well under smooth motion but often degrade in fast dynamics or low-light environments.

\noindent \textbf{Deep learning-based methods}. 
Recent advances in deep learning have enabled data-driven event denoising through end-to-end learning. These models are typically trained on noisy-clean event pairs or self-supervised proxies. Early work like K-SVD~\cite{xie2018dvs} employed sparse feature learning, followed by EDnCNN~\cite{baldwin2020event}, which fuses frame and IMU data via convolutional networks. EventZoom~\cite{duan2021eventzoom} introduced noise-to-noise training with a U-Net, and AEDNet~\cite{fang2022aednet} leveraged PointNet to process raw event streams. MLPF~\cite{guo2022low} explored probabilistic modeling, while Alkendi et al.\cite{alkendi2022neuromorphic} combined GNNs and transformers for per-event classification. EDformer\cite{jiang2024edformer} adopts a pure transformer architecture for event-wise denoising. Although not designed for denoising, Pre-Mamba~\cite{ruan2025pre} extends state space models to 4D event sequences for deraining, highlighting their potential for high-resolution event modeling.

\begin{figure*}[t]
  \centering
  \includegraphics[width=0.98\linewidth]{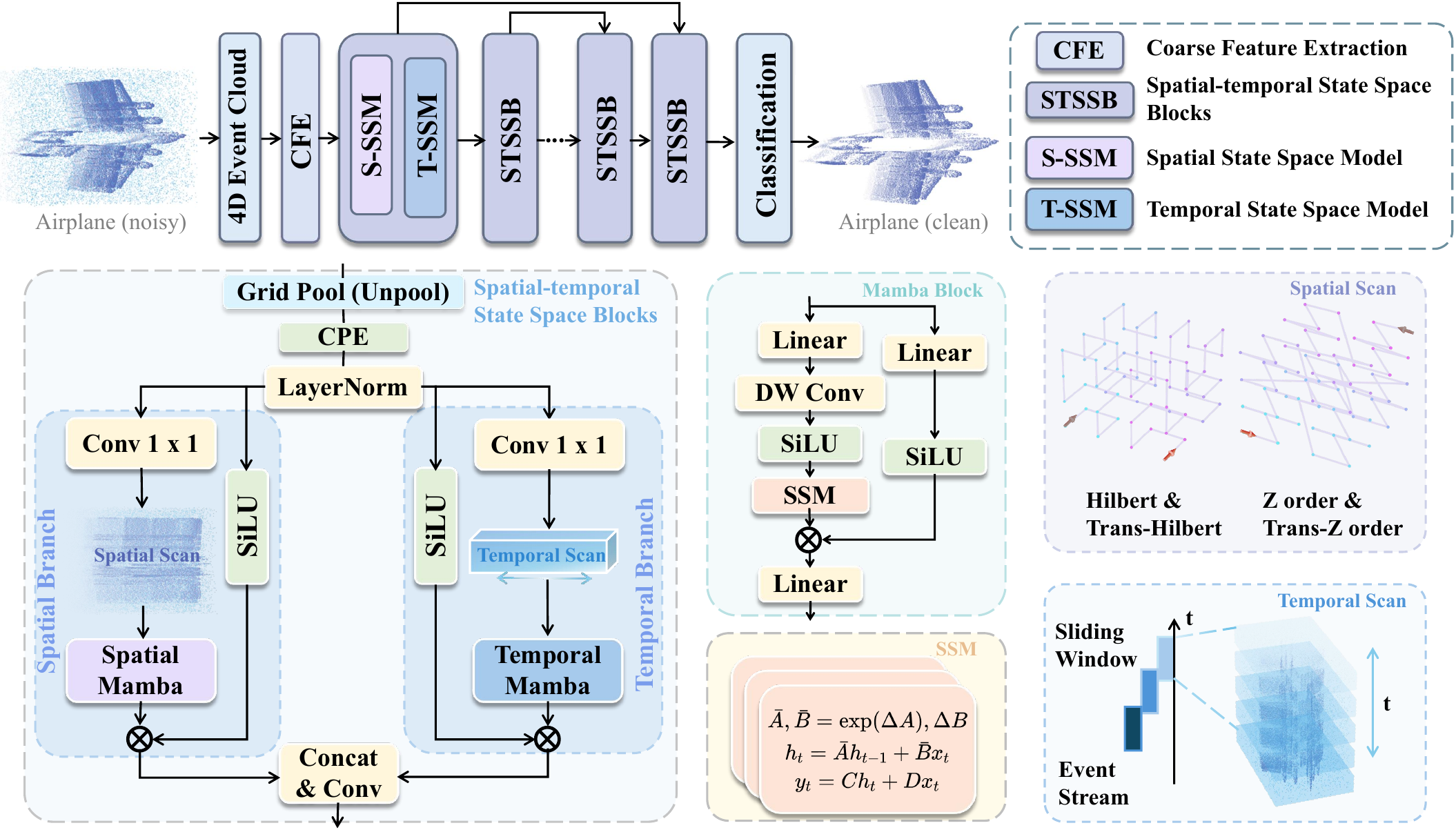}
  % \fbox{\rule{0pt}{4in} \rule{0.9\linewidth}{0pt}}
  \vspace{-0.2cm}
  \caption{Overview of the EDmamba architecture. Raw events are grouped into 4D event clouds capturing spatial, temporal, and polarity information. The Coarse Feature Extraction (CFE) module projects events into geometric and polarity-aware subspaces. The U-Net-style denoising backbone employs multi-scale Spatial-Temporal State Space Blocks (STSSBs), composed of two complementary modules: Spatial SSM (S-SSM) suppresses spatially incoherent noise by modeling local geometric consistency, while Temporal SSM (T-SSM) filters temporally inconsistent events by capturing motion-aligned patterns across time. }
  \vspace{-0.3cm}
  \label{fig:overview}
\end{figure*}

\section{Method}
\label{sec:method}
\subsection{Working Principle}

Our event denoising framework is grounded in the physical mechanisms of event generation and sensor noise.  
An event is triggered at pixel $\mathbf{u} = (x, y)^\top$ when the log-intensity change exceeds a contrast threshold $C$, yielding a polarity $p \in \{-1, +1\}$:
\begin{equation} \label{eq:event_trigger}
    pC = \Delta L(\mathbf{u},t) \triangleq \log I(\mathbf{u},t) - \log I(\mathbf{u},t-\Delta t).
\end{equation}
This change can be decomposed as $\Delta L = \Delta L_s + \Delta L_n + \Delta L_c$, where $\Delta L_s$ denotes motion-induced signal, approximated by:
\begin{equation} \label{eq:signal}
\Delta L_s \approx -\nabla \log I \cdot \mathbf{v} \Delta t,
\end{equation}
where $\nabla \log I$ is the spatial log-intensity gradient and $\mathbf{v}$ the image-plane velocity.
% The remaining terms $\Delta L_n$ and $\Delta L_c$ denote photonic and circuit-level noise. $\Delta L_n$ models photon shot noise and background activity, leading to temporally incoherent events; $\Delta L_c$ reflects thermal leakage and fixed-pattern effects that cause spatially inconsistent activations:
The remaining terms $\Delta L_n$ and $\Delta L_c$ denote photonic and circuit-level noise, respectively. Specifically, $\Delta L_n$ models photon shot noise and background activity, often causing temporally incoherent firings, while $\Delta L_c$ captures thermal leakage and fixed-pattern artifacts such as hot pixels, leading to spatially inconsistent activations:
\begin{equation} \label{eq:noise}
\begin{aligned}
\Delta L_n &\sim \mathcal{N}(0,\sigma_n^2) + \lambda_{BA} \mathcal{P}(\gamma_{BA}), \\
\Delta L_c &= \eta_{th}\left(\frac{k_B T}{e}\right) + \frac{I_{dark} \Delta t}{C_{pd}},
\end{aligned}
\end{equation}
where $\sigma_n$ is the readout noise, $\lambda_{BA}$ the background activity rate, $\gamma_{BA}$ the gain, $k_B$ the Boltzmann constant, $T$ the temperature, $e$ the electron charge, $I_{dark}$ the leakage current, and $C_{pd}$ the photodiode capacitance.

These noise sources manifest in distinct ways on the event stream. As visualized in Fig.~\ref{fig:feature}(a), signal events exhibit smooth, motion-aligned trajectories. In contrast, background activity introduces jittery temporal spikes that disrupt coherence, while hot pixels continuously fire at fixed positions, violating spatial consistency. To differentiate such patterns, we implement a state-space classifier that integrates local spatial and temporal neighborhoods:
% As visualized in Fig.~\ref{fig:feature}(a), signal events exhibit smooth motion-aligned trajectories, while background activity introduces temporally incoherent firings, and hot pixels persistently activate at fixed locations, violating spatial consistency. We implement this via a state-space classifier that integrates local spatial and temporal neighborhoods:
\begin{equation} \label{eq:classifier}
f_\theta: (\mathcal{N}_s(e_i), \mathcal{N}_t(e_i)) \rightarrow \{0,1\},
\end{equation}
where $\mathcal{N}_s$ and $\mathcal{N}_t$ denote spatial and temporal neighborhoods. The classifier integrates local geometry and motion continuity to suppress structured spatial and temporal noise.

\subsection{EDmamba}
\noindent \textbf{Overview Architecture.} 
Fig.~\ref{fig:overview} illustrates the architecture of \textbf{EDmamba}, a dual-branch encoder-decoder framework for event denoising. Raw events are encoded as a spatiotemporal point cloud and structured into a 4D tensor. A Coarse Feature Extraction (CFE) stage applies depthwise convolutions and linear projections to jointly encode geometric and polarity-aware features. To address different noise types, EDmamba includes two decoupled branches: the \textbf{Spatial SSM (S-SSM)} targets location-dependent structured noise such as leakage and fixed-pattern effects, while the \textbf{Temporal SSM (T-SSM)} handles temporally uncorrelated background activity modeled as Poisson and thermal noise. For directional modeling, the 4D event cloud is flattened into three sequences: two spatial (via space-filling curves) and one temporal (via time scan). These are processed through a U-Net-style hierarchy of multi-scale Spatial-Temporal State Space Blocks (STSSBs), which embed S-SSM and T-SSM modules to model spatial and temporal dependencies. Fused features are decoded to generate denoised events, with skip connections preserving information across scales.

\begin{figure}[t]
  \centering
  \includegraphics[width=\linewidth]{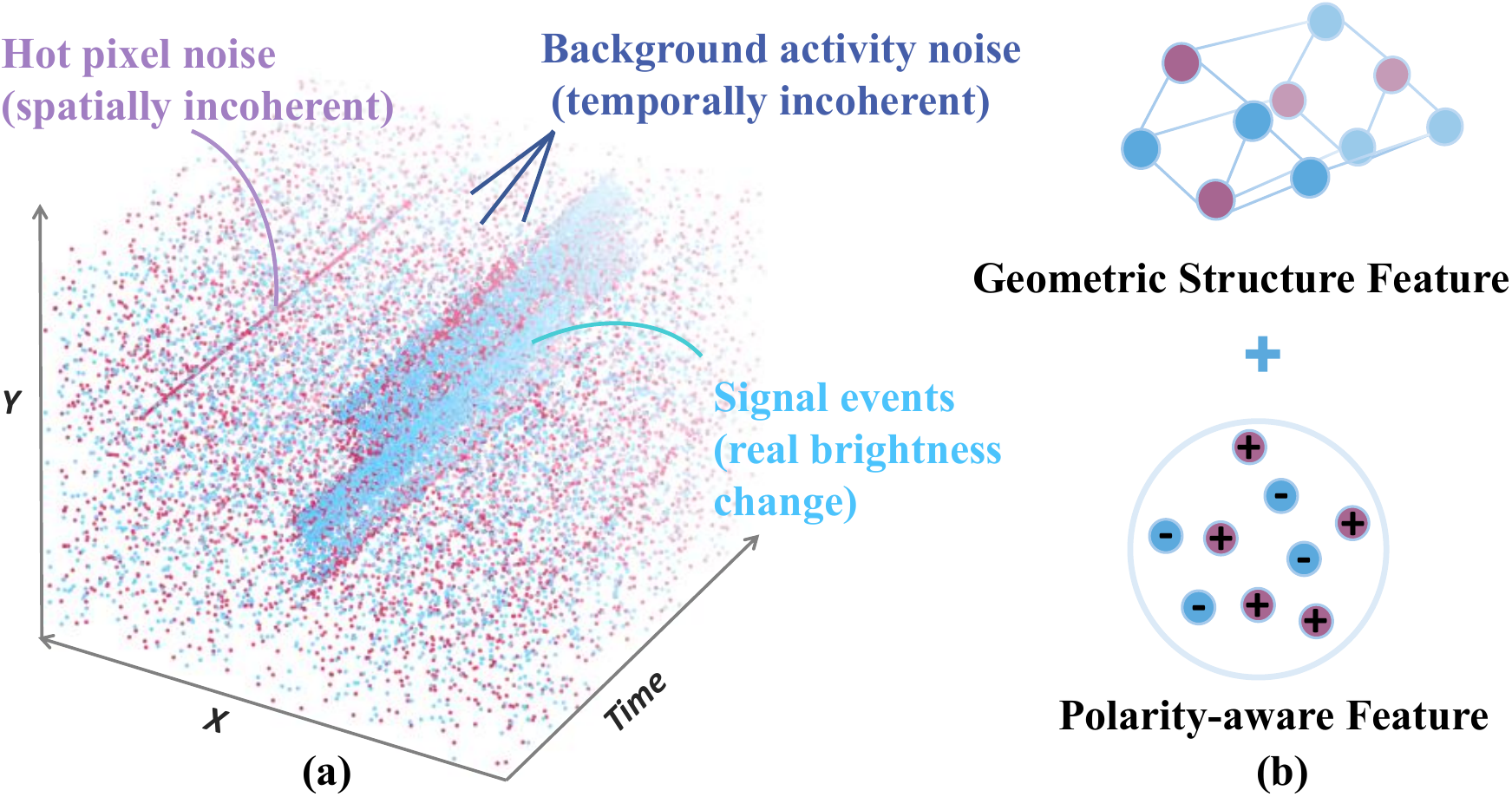}
    \vspace{-0.5cm}
  \caption{(a) Visualization of signal events and spatiotemporally incoherent noise. (b) Feature decomposition into geometry and polarity components in CFE module. Red and blue dots indicate ON and OFF polarity events, respectively.}
  \label{fig:feature}
  \vspace{-0.6cm}
\end{figure}

\noindent \textbf{Input Representation.} 
We represent raw events as $E = \{ e_i \}_{i=1}^N$, where each $e_i = (x_i, y_i, t_i, p_i)$ denotes an event with spatial coordinates $(x_i, y_i)$, timestamp $t_i$, and polarity $p_i \in \{-1, 1\}$. The event stream is divided into $L$ consecutive segments $\{S_k\}_{k=1}^L$, each containing $N$ events. Within each segment $S_k$, timestamps are normalized using the first and last event times, $t_0 = t_{\text{start}}^k$ and $t_e = t_{\text{end}}^k$, as:
\vspace{-0.1cm}
\begin{equation}
    z_i = \frac{t_i - t_0}{t_e - t_0}, \quad \text{for } e_i \in S_k,
\end{equation}
yielding a 3D pseudo point cloud $(x_i, y_i, z_i)$. Incorporating polarity $p_i$ gives a 4D event cloud $(x_i, y_i, z_i, p_i)$ that preserves spatial coordinates, temporal order, and polarity cues. This representation preserves spatial structure, captures local temporal context via normalized timestamps, and retains polarity information in a compact, learnable form. 

% This representation avoids the artifacts commonly introduced by event stacking methods.
% we obtain a 4D event cloud representation $(x_i, y_i, z_i, p_i)$ that preserves both spatial detail, temporal continuity and polarity information. This representation maintains both spatial detail and  temporal continuity, avoiding the artifacts often introduced by event stacking approaches.

\noindent \textbf{Event Sampling.}
To reduce redundancy while retaining salient motion patterns, we apply structured sampling to the high-rate event stream. Events are voxelized by discretizing the normalized timestamp $z_i$ with voxel size $v$, forming integer coordinates $(x_i, y_i, \lfloor z_i/v \rfloor, p_i)$. A spatial hash function $\mathcal{H}: \mathbb{Z}^4 \rightarrow \mathbb{N}$ assigns each voxel a unique key. During training, one event is randomly sampled from each non-empty voxel, preserving statistical diversity while reducing computation and suppressing spurious noise.

\noindent \textbf{Coarse Feature Extraction.}
To support effective denoising, we propose a Coarse Feature Extraction (CFE) module that decomposes the 4D event cloud into two modality-specific components: geometric structure and polarity signal (Fig.~\ref{fig:feature}(b)). This decomposition reflects two fundamental properties of event data: \textit{(i)} signal events often form coherent geometric patterns aligned with motion; \textit{(ii)} polarity provides an additional modality that is unique to event cameras, and polarity inconsistencies are frequently indicative of noise such as flip errors or unstructured background firing.

Given a segment of events $\mathcal{E}_n = \{(x_i, y_i, z_i, p_i)\}_{i=1}^{M_n}$, CFE applies  1D convolutions with activation functions to extract axis-wise features. 
The geometric and polarity branches operate on $(x, y, z)$ and $(x, y, p)$ respectively, encoding motion-aligned structures and polarity consistency. Axis-wise convolutions extract modality-specific features, which are subsequently fused via a $1 \times 1$ projection:
\begin{equation}
f_i = \text{Conv}_{1\times1}\{(\phi_\text{geom}, \phi_\text{pol})_{\text{concat}}\},
\end{equation}
yielding a compact point-wise embedding $f_i$ that preserves spatial, temporal, and polarity-aware cues. These features are subsequently passed to two decoupled branches, each specialized in suppressing either structured spatial artifacts or stochastic temporal noise.

% The geometric branch processes $(x_i, y_i, z_i)$, where the normalized timestamp $z$ implicitly encodes local temporal cues, enabling the branch to capture motion-aligned spatiotemporal structures.  In parallel, the polarity branch models the spatial distribution of ON/OFF transitions using $(x_i, y_i, p_i)$. The two feature streams are concatenated and fused via a $1\times1$ point-wise convolution:
% \begin{equation}
% f_i = \text{Conv}_{1\times1}\{(\phi_\text{geom}, \phi_\text{pol})_{\text{concat}}\},
% \end{equation}
% resulting in a compact point-wise embedding $f_i$ that preserves spatial, temporal, and polarity-specific cues.

% \begin{figure}[t]
%   \centering
%   \includegraphics[width=\linewidth]{img/2_feature.pdf}
%     \vspace{-0.5cm}
%   \caption{Coarse-Level Decomposition of Event Cloud into Geometric and Polarity Features.}
%   \label{fig:feature}
%   \vspace{-0.7cm}
% \end{figure}
\begin{table*}[htbp]
\setlength{\tabcolsep}{2.5mm}
\fontsize{9}{10}\selectfont
\centering
\begin{tabular}{ccccccccccc}
\hline
\multicolumn{11}{c}{DND21 (346 $\times$ 260)}                                                                                                                                                                      \\ \hline
\multirow{2}{*}{Methods} & \multicolumn{2}{c}{1 Hz/pixel} & \multicolumn{2}{c}{3 Hz/pixel} & \multicolumn{2}{c}{5 Hz/pixel} & \multicolumn{2}{c}{7 Hz/pixel} & \multicolumn{2}{c}{10 Hz/pixel} \\ \cline{2-11} 
                         & Hotel-bar       & Driving      & Hotel-bar       & Driving      & Hotel-bar       & Driving      & Hotel-bar       & Driving      & Hotel-bar       & Driving       \\ \hline
EvFlow                   & 0.5137               & 0.5341            & 0.5144               & 0.5351            & 0.5154               & 0.5359            & 0.5160               & 0.5368            & 0.5172               & 0.5379             \\
TS                      & 0.5785               & 0.6824            & 0.5721               & 0.6262           & 0.5695               & 0.6250            & 0.5688               & 0.6249            & 0.5689               & 0.6246             \\
DWF                     & 0.8911               & 0.6592            & 0.8616               & 0.6532           & 0.8778               & 0.6502            & 0.8683               & 0.6452            & 0.8563               & 0.6366             \\
KNoise                   & 0.6780               & 0.6297            & 0.6524               & 0.6203            & 0.6579               & 0.6201            & 0.6489               & 0.6148            & 0.6413               & 0.6146             \\
YNoise                  & 0.7699               & 0.8086            & 0.7658               & 0.8041            & 0.7594               & 0.7978            & 0.7553               & 0.7949            & 0.7507               & 0.7873            \\
RED                      & 0.6475               & 0.5873           & 0.6571               & 0.5913            & 0.6634               & 0.5944            & 0.6721               & 0.6003           & 0.6867               & 0.6062             \\
AEDNet                  & 0.8070               & 0.8368            & 0.8568               & 0.8337            & 0.9561               & 0.8325            & 0.8850               & 0.8071            & 0.8990               & 0.8201             \\
EDnCNN     &     0.9573 & 0.8873 & 0.9371 & 0.8771 & 0.9365 & 0.8748 & 0.9254 & 0.8654&  0.9006 & 0.8574           \\
EDformer               & \underline{0.9928}               & \underline{0.9542}            & \underline{0.9891}               & \underline{0.9472}            & \underline{0.9845}               & \underline{0.9424}            & \underline{0.9793 }             & \underline{0.9344}            &\underline{0.9699}               & \underline{0.9264}            \\
Ours             & \textbf{0.9963}               & \textbf{0.9694}            & \textbf{0.9949 }              & \textbf{0.9710}            & \textbf{0.9955}               & \textbf{0.9734}            &\textbf{ 0.9939}               & \textbf{0.9689}            & \textbf{0.9916}               &      \textbf{ 0.9636}              \\ \hline
\multicolumn{11}{c}{DVSCLEAN (1280 $\times$ 720)}                                                                                                                                                                  \\ \hline
\multirow{2}{*}{Methods} & \multicolumn{2}{c}{Double-bracket}    & \multicolumn{2}{c}{Double-ship}    & \multicolumn{2}{c}{Double-airplane}    & \multicolumn{2}{c}{ Multi-helicopter
}    & \multicolumn{2}{c}{ Multi-car}     \\ \cline{2-11} 
                         & 50\%            & 100\%        & 50\%            & 100\%        & 50\%            & 100\%        & 50\%            & 100\%        & 50\%            & 100\%         \\ \hline
EvFlow                  & 0.6221               & 0.7591            & 0.6998               & 0.6959            & 0.8100               & 0.7922            & 0.8562               & \underline{0.8368}            & 0.7919               & 0.7808             \\
TS                    & 0.8303               & 0.8092            & 0.8054               & 0.7919            & 0.8507               & \underline{0.8120}            & 0.8749               & 0.8343            & 0.8707               & \underline{0.8438}             \\
DWF                   & 0.5995               & 0.6313            & 0.5998               & 0.6325            & 0.5770               & 0.5954            & 0.6201               & 0.6048            & 0.6098               & 0.5957             \\
KNoise               & 0.5958               & 0.5765            & 0.5950               & 0.5805            & 0.5751               & 0.5563            & 0.5909               & 0.5713            & 0.5975               & 0.5775             \\
YNoise                 & 0.6194               & 0.6156            & 0.5903               & 0.5886            & 0.6634               & 0.6529            & 0.7504               & 0.7356            & 0.6540               & 0.6471             \\
RED                 & 0.5972               & 0.6792            & 0.6469               & 0.6439            & 0.7357               & 0.7109            & 0.7308               & 0.7062            & 0.7458               & 0.7289             \\
AEDNet                 & 0.7314               & 0.6349            & 0.7142               & 0.6450            & 0.5530               & 0.5158            & 0.6509               & 0.5822            & 0.6160               & 0.5409             \\
EDnCNN                 & \underline{0.9432}               & 0.7766           & \underline{0.9473}               & 0.7776            & \underline{0.9295}              & 0.7827            & \textbf{0.9394}               & 0.7679            & \textbf{0.9301}               & 0.7615             \\

EDformer             & 0.9209               & \underline{0.8565}            & 0.9382               & \underline{0.8801}            & 0.8745               & 0.7924            & 0.8945               & 0.8251           & 0.9056               & 0.8431            \\
Ours           & \textbf{0.9457}               & \textbf{0.9248}           & \textbf{0.9684}               & \textbf{0.9572}            & \textbf{0.9684}               & \textbf{0.8704}            & \underline{0.9247}               & \textbf{0.9122}            & \underline{0.9218}               & \textbf{0.9036}                       \\ \hline
\end{tabular}

% \vspace{-0.2cm}
% \caption{The AUC of different denoising methods on the test set of DND21 and DVSCLEAN at different shot noise rates. We mark the \textbf{best} and \underline{second best} results.}
\caption{AUC results on DND21 and DVSCLEAN under varying shot noise rates. \textbf{Best} and \underline{second best} results are highlighted.}
\vspace{-0.6cm}
\label{tab:auc}
\end{table*}

% \vspace{-0.1cm}
\subsection{S-SSM: Spatial Modeling with Local Geometric Priors.}
Spatial noise in event cameras often originates from fixed-pattern leakage, hot pixels, or circuit-level inconsistencies. These recurring artifacts usually exhibit location-dependent repetition and significantly disrupt the surrounding local geometric coherence. To mitigate such structured artifacts, we design a Spatial State-Space Module (S-SSM) that explicitly incorporates spatial priors.

S-SSM leverages space-filling curves (e.g., Z-order, Hilbert) to flatten the 3D spatial domain into sequences while preserving neighborhood continuity. This allows the model to reason over local geometric patterns and edge structures. The sequences are processed by Mamba blocks with depthwise convolutions and bidirectional state updates, capturing both short- and long-range dependencies efficiently. The design biases the model toward spatial smoothness, enabling it to detect structural edges and suppress isolated or repetitive noise. As spatial noise lacks meaningful scene-driven causes, S-SSM focuses on enforcing geometric regularity rather than modeling causality.
% \vspace{-0.1cm}
\subsection{T-SSM: Temporal Modeling with Motion Continuity.}
Temporal noise, caused by shot noise, thermal fluctuations, or background activity, severely disrupts the consistency of event streams due to its highly random and unstructured temporal nature. To address this challenge, we introduce the Temporal State-Space Module (T-SSM), which accurately captures motion-consistent patterns by explicitly modeling bidirectional temporal dependencies.

T-SSM first sorts events by normalized timestamp to form a temporally ordered sequence, which is processed by a bidirectional Mamba block to capture forward and backward motion patterns. By learning global temporal consistency, T-SSM suppresses scattered or flickering events while preserving coherent trajectories. This design reflects a key physical prior: real motion yields causally consistent patterns, while temporal noise is fundamentally acausal. To exploit this distinction, T-SSM learns consistent transitions to remove incoherent activations, working alongside the spatial branch for joint spatiotemporal denoising.

\vspace{-0.2cm}
\section{Experiments}
\label{sec:experiment}
\begin{figure*}[t]
  \centering
  \includegraphics[width=0.99\linewidth]{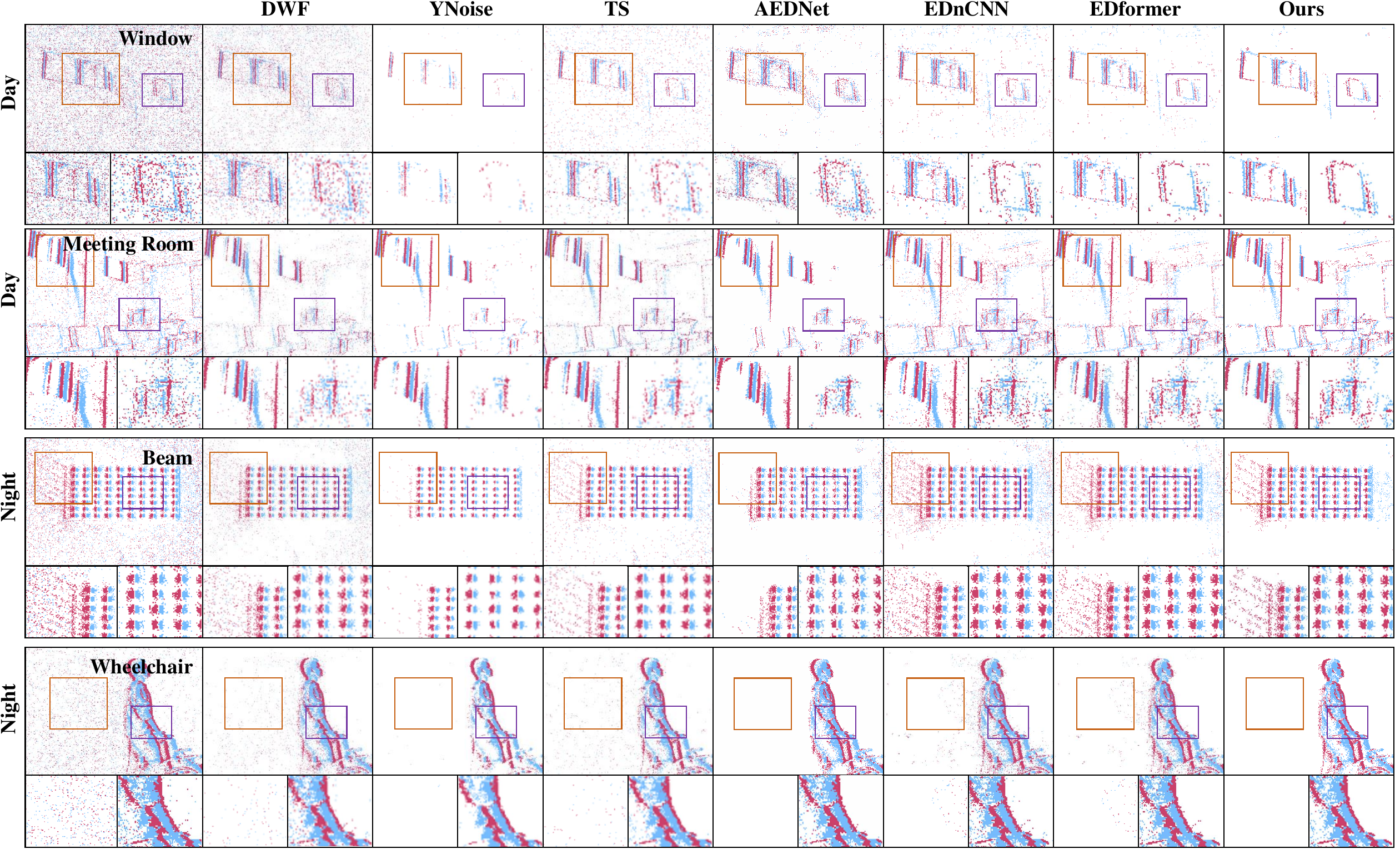}
  % \fbox{\rule{0pt}{4.2in} \rule{0.9\linewidth}{0pt}}
  \vspace{-0.25cm}
  \caption{Visual comparison on the  E-MLB~\cite{ding2023mlb} dataset under daytime (top two rows) and nighttime (bottom two rows) conditions. EDmamba effectively suppresses background noise while preserving fine motion and structural details. In contrast, baseline methods either leave residual noise or blur object contours, especially under low-light conditions.}
  \label{fig:qualitative2}
  % \vspace{-0.55cm}
\end{figure*}
% \begin{figure*}[t]
%   \centering
%   \includegraphics[width=\linewidth]{img/3_result1.pdf}
%   % \fbox{\rule{0pt}{4in} \rule{0.9\linewidth}{0pt}}
%   \caption{Visual comparison on the DND21 dataset under 5Hz/pixel and 10Hz/pixel noise levels in hotel-bar and driving scenes.}
%   \label{fig:qualitative1}
% \end{figure*}
% \subsection{Experiments Setups}

\noindent\textbf{Implementation Details.}
We optimize EDmamba using a cross-entropy objective, following standard practices in event denoising~\cite{fang2022aednet, jiang2024edformer}. The model is implemented in PyTorch and trained for 50 epochs on eight NVIDIA RTX A6000 GPUs with a batch size of 128. We use the AdamW optimizer with an initial learning rate of $8\times10^{-5}$ per sample and a weight decay of $5\times10^{-2}$ to ensure stable convergence and effective regularization.
During training, input events are voxelized along the $z$-axis with a grid resolution of 0.1 and a fixed sample size $N=10240$, preserving the spatiotemporal structure while maintaining computational efficiency.
The network adopts a U-Net-style architecture~\cite{ronneberger2015u}, consisting of a two-stage encoder and a single-stage decoder with block depths of [2, 4] and [2], respectively. The encoder applies serialized pooling (scale factor 2) after the first stage, increasing channels from 8 to 16. The decoder uses serialized unpooling and skip connections for multi-scale fusion.

\noindent\textbf{Datasets.}
% We evaluate EDmamba across diverse event denoising datasets spanning both synthetic and real-world settings. For supervised training, we adopt ED24~\cite{jiang2024edformer}, a large-scale dataset targeting background activity (BA) noise, recorded under 21 illumination levels using DAVIS346 with optical instrumentation for fine-grained noise characterization. 
% Evaluation is conducted on two labeled benchmarks: DVSCLEAN~\cite{fang2022aednet}, a synthetic dataset generated via ESIM with injected noise, and DND21~\cite{guo2022low}, a v2e-simulated dataset with limited size. To assess generalization, we further test on the unlabeled real-world dataset E-MLB~\cite{ding2023mlb}, which features diverse indoor and outdoor motions under varying lighting, recorded with DAVIS346 and aligned APS/IMU data. 
% This setup enables comprehensive evaluation on both benchmark performance and real-world robustness.
We evaluate EDmamba on both labeled and unlabeled event datasets across synthetic and real-world domains. Supervised training is conducted on ED24~\cite{jiang2024edformer}, a large-scale dataset annotated for background activity noise, collected under 21 controlled illumination levels with DAVIS346.
Quantitative evaluation is performed on two labeled benchmarks: DVSCLEAN~\cite{fang2022aednet}, a synthetic dataset generated via ESIM with noise injection, and DND21~\cite{guo2022low}, a v2e-based dataset with frame-level supervision.
To assess generalization and denoising quality, we further evaluate on unlabeled real-world datasets E-MLB~\cite{ding2023mlb}, which contains diverse indoor and outdoor motions under varying lighting.
% To assess real-world generalization, we test on two unlabeled datasets: E-MLB~\cite{ding2023mlb}, which captures diverse indoor and outdoor motions under varying lighting, and RGBDAVIS, a multi-modal dataset with synchronized RGB and event streams.
% This setup enables comprehensive evaluation across denoising accuracy, generalization, and robustness to real sensor noise.

\begin{table*}[htbp]
\fontsize{9}{10.3}\selectfont
\begin{threeparttable}
\begin{tabular*}{\textwidth}{@{\extracolsep{\fill}}lccccccccccc}
\toprule
\multicolumn{1}{c}{\multirow{2}{*}{Method}}  & \multicolumn{4}{c}{E-MLB (Daylight)} & \multicolumn{4}{c}{E-MLB (Night)}  & DND21 \\ \cline{2-10} 
                        & ND1     & ND4    & ND16    & ND64    & ND1    & ND4    & ND16   & ND64   & -          \\ \hline
Raw                                              & 0.821 & 0.824 & 0.815 & 0.786 & 0.890 & 0.824 & 0.786 & 0.768 & 0.869 \\
TS ~\cite{lagorce2016hots}                      & 0.943 & 0.955 & 0.980 & \textbf{0.995} & 0.938 & 0.884 & 0.859 & 0.907 & 0.954 \\
EvFlow ~\cite{wang2019ev}                       & 0.871 & 0.919 & 0.917 & 0.921 & 0.951 & 0.876 & 0.852 & 0.886 & 1.034 \\
IETS $^{\dagger}$ ~\cite{baldwin2019inceptive}  & 0.772 & 0.785 & 0.777 & 0.753 & 0.950 & 0.823 & 0.804 & 0.711 & 0.900 \\
KNoise ~\cite{khodamoradi2018n}                & 0.787 & 0.819 & 0.810 & 0.786 & 0.904 & 0.842 & 0.819 & 0.860 & 0.998 \\
EDnCNN ~\cite{baldwin2020event}                & 0.887 & 0.908 & 0.903 & 0.912 & 1.001 & \underline{1.024} & \underline{1.079} & 1.086 & 0.977 \\
YNoise ~\cite{Feng2020EventDB}                 & 0.902 & 0.922 & 0.917 & 0.934 & 0.962 & 0.895 & 0.874 & 0.928 & 0.984 \\
GET $^{\dagger}$ ~\cite{duan2021guided}        & \textbf{1.051} & 0.938 & 0.935 & 0.927 & 1.027 & 0.955 & 0.946 & 0.935 & 0.932 \\
EventZoom $^{\dagger}$ ~\cite{duan2021eventzoom} & \underline{0.996} & \underline{0.988} & \textbf{0.996} & 0.970 & \textbf{1.055} & 1.007 & 1.010 & 0.988 & \textbf{1.059} \\
MLPF ~\cite{guo2022low}                         & 0.851 & 0.855 & 0.846 & 0.840 & 0.926 & 0.928 & 0.910 & 0.906 & 0.944 \\
DWF ~\cite{guo2022low}                          & 0.932 & 0.945 & 0.943 & 0.904 & 0.916 & 0.871 & 0.825 & 0.873 & 0.972 \\
AEDNet ~\cite{fang2022aednet}                  & 0.789 & 0.836 & 0.803 & 0.789 & 0.887 & 0.929 & 0.929 & 0.958 & 0.919 \\
RED ~\cite{ding2023mlb}                         & 0.971 & 0.943 & 0.946 & 0.923 & 0.948 & 0.973 & 1.001 & 0.916 & 0.945 \\
EDformer ~\cite{jiang2024edformer}             & 0.952 & 0.955 & 0.956 & 0.942 & \underline{1.048} & 1.019 & 1.076 & \textbf{1.099} & 1.041 \\
Ours                                             & 0.976 & \textbf{0.990} & \underline{0.985} & \underline{0.972} & 1.002 & \textbf{1.025} & \textbf{1.082} & \underline{1.089} & \underline{1.057} \\
\bottomrule
\end{tabular*}
\begin{tablenotes}
% \item nan: ESR not applicable due to insufficient event density (<5\% events retained).
\item $^{\dagger}$: The result is derived from E-MLB~\cite{ding2023mlb}, as the official code is not publicly available.
\end{tablenotes}
\end{threeparttable}
% \vspace{-0.3cm}
% \caption{The mean ESR (MESR) results of different denoising methods on E-MLB and  DND21 event denoising datasets. We mark the \textbf{best} and \underline{second best} results.}
\caption{The MESR results of denoising methods on E-MLB and DND21 datasets. \textbf{Best} and \underline{second-best} values are highlighted.}
\vspace{-0.5cm}
\label{tab:mesr}
\end{table*}

\begin{table}[htbp]
\centering
\fontsize{9}{10}\selectfont
\setlength{\tabcolsep}{1.5mm}  % 压缩列间距
\begin{tabular}{ccccc}
\toprule
Method & GFLOPs & \#Params & Inf. Time (s) & Rel. Speed \\
\midrule
TS        & N/A     & N/A     & 0.1296  & 1.0$\times$ \\
EvFlow    & N/A     & N/A     & 1.5545  & 0.08$\times$ \\
DWF       & N/A     & N/A     & 0.0954  & 1.36$\times$ \\
KNoise    & N/A     & N/A     & \textbf{0.0198} & \textbf{6.55}$\times$ \\
YNoise    & N/A     & N/A     & \underline{0.0513} & \underline{2.53}$\times$ \\
RED       & N/A     & N/A     & 2.2716  & 0.06$\times$ \\
\midrule
EDnCNN    & 234.51  & 614.55K & 20.1885 & 1.0$\times$ \\
AEDNet    & 4400.46 & 45.87M  & 43.4250 & 0.46$\times$ \\
EDformer  & 8.41 & \textbf{49.80K} & 2.4943 & 8.09$\times$ \\
Pre-Mamba$^{\dagger}$& 6.23 & 264.63K & 0.0987 & 204.54$\times$\\
Joint-SSM&\underline{5.17}&102.91K&\underline{0.0931}&\underline{216.85$\times$} \\
Ours      & \textbf{2.27} & \underline{88.98K} & \textbf{0.0685} & \textbf{294.72}$\times$ \\
\bottomrule
\end{tabular}
\begin{tablenotes}
% \item nan: ESR not applicable due to insufficient event density (<5\% events retained).
\item $^{\dagger}$: The result is derived from Pre-Mamba~\cite{ruan2025pre}.
\end{tablenotes}
% \vspace{-0.3cm}
\caption{Efficiency comparison in terms of GFLOPs, parameters, inference time, and relative speed (100K events).}
\label{tab:efficiency}
\vspace{-0.6cm}
\end{table}

\noindent\textbf{Compared Methods.}
We conduct extensive comparisons with state-of-the-art event denoising methods across both traditional and learning-based categories. For conventional approaches, we evaluate against density-based filters: BAF~\cite{delbruck2008frame}, KNoise~\cite{khodamoradi2018n}, DWF~\cite{guo2022low}, and YNoise~\cite{Feng2020EventDB}; time-surface methods: TS~\cite{lagorce2016hots} and IETS~\cite{baldwin2019inceptive}; the recursive event denoiser MLB~\cite{ding2023mlb}; optical flow-based method EvFlow~\cite{wang2019ev}; and the guided filter GEF~\cite{duan2021guided}. 
On the learning-based side, we compare with MLP-based method MLPF~\cite{guo2022low}, CNN-based models EDnCNN~\cite{baldwin2020event} and EventZoom~\cite{duan2021eventzoom}, the PointNet-based AEDNet~\cite{fang2022aednet}, and the Transformer-based EDformer~\cite{jiang2024edformer}.

\noindent\textbf{Metrics.}
For labeled datasets (DVSCLEAN~\cite{fang2022aednet} and DND21~\cite{guo2022low}), we evaluate denoising performance using the Area Under the Curve (AUC) of event-level predictions, computed from binary ground-truth labels. For the unlabeled dataset E-MLB~\cite{ding2023mlb}, we adopt the Mean Event Structural Ratio (MESR)~\cite{ding2023mlb}, which quantifies structural consistency by measuring contrast enhancement in motion-compensated event volumes. Unlike label-dependent metrics, MESR leverages statistical regularities in the event stream and does not require annotations or auxiliary modalities, making it suitable for real-world evaluation.
% For completeness, we also report MESR on DND21 to enable unified comparisons across labeled and unlabeled datasets.
% Additional metrics including Signal Retention (SR), Noise Removal (NR), and Denoising Accuracy (DA) are provided in the supplementary for a more detailed evaluation on labeled data.
\vspace{-0.2cm}
\subsection{Quantitative Evaluation}
% To evaluate denoising performance on labeled datasets, we compute AUC scores following the evaluation setting in~\cite{guo2022low}. The results on the test sets of DND21~\cite{guo2022low} and DVSCLEAN~\cite{fang2022aednet} are reported in Tab.~\ref{tab:auc}. DND21~\cite{guo2022low} includes two test scenes with v2e-simulated shot noise (1–10 Hz/pixel), while DVSCLEAN~\cite{fang2022aednet} provides five test scenes under two noise levels: 50\% and 100\% noise ratio. 
% % The corresponding ROC curves are provided in the supplementary. 
% Our EDmamba model consistently achieves the highest AUC across both datasets, demonstrating robust discrimination between signal and noise under varying event statistics and noise levels.

To evaluate denoising performance on labeled datasets, we compute AUC scores following the evaluation protocol in~\cite{guo2022low}, with results reported in Tab.~\ref{tab:auc}. DND21 includes two 346×260 test scenes with shot noise rates from 1–10 Hz/pixel, simulating low-light conditions. DVSCLEAN provides five 1280×720 sequences under two noise levels (50\% and 100\%).
Our EDmamba achieves the highest AUC scores on DND21 by effectively handling varying shot noise levels. On average, EDmamba achieves AUC scores of 0.9944 and 0.9693 on the hotel-bar and driving scenes, outperforming EDformer with relative AUC improvements of 1.15\% and 3.01\%.
While EDmamba achieves leading performance on most DVSCLEAN sequences, EDnCNN slightly outperforms it in a few cases due to its use of $k$-nearest spatio-temporal neighbors for fine-grained event aggregation. However, this explicit neighbor search increases computational cost (Tab.~\ref{tab:efficiency}), whereas EDmamba processes raw streams directly with higher efficiency.
% and outperforming all competing methods at both noise levels in DVSCLEAN.
% In addition, EDmamba achieves a \todo{X\% and Y\%} higher AUC score compared to EDformer on DND21~\cite{guo2022low} and DVSCLEAN dataset~\cite{fang2022aednet}, respectively. 
% This robust performance highlights EDmamba's ability to adapt to different sensor resolutions and noise characteristics while maintaining exceptional signal-noise discrimination.

% To further evaluate the generalization of EDmamba in label-free, real-world scenarios, we conduct MESR testing on the E-MLB (Daylight, Night) and RGBDAVIS (Indoor, Outdoor) datasets.  As shown in Tab.~\ref{tab:mesr}, EDmamba consistently achieves high MESR scores, demonstrating stable denoising performance even in low-light conditions. We also compute MESR on DND21, further supporting our method's generalization across synthetic and real-world datasets.
% To further evaluate the generalization of EDmamba in label-free, real-world scenarios, we conduct MESR testing on the E-MLB (Daylight, Night) and DND21 datasets. As shown in Tab.~\ref{tab:mesr}, EDmamba consistently delivers strong MESR performance, indicating robust denoising even under challenging lighting. While methods like YNoise and TS tend to report high MESR by indiscriminately removing both noise and informative background content, leading to over-denoising, EDmamba achieves a better balance between noise suppression and content preservation.
To further evaluate the generalization of EDmamba in label-free, real-world scenarios, we conduct MESR testing on the E-MLB (Daylight, Night) and DND21 datasets. As shown in Tab.~\ref{tab:mesr}, EDmamba consistently delivers strong MESR performance, demonstrating robust denoising under challenging lighting. While methods like TS tend to achieve high MESR by over-suppressing both noise and informative background content, EDmamba strikes a better balance between denoising and content preservation.

\vspace{-0.1cm}
\subsection{Qualitative Evaluation}
To illustrate the effectiveness of our method, we present visual comparisons across different datasets and noise levels. 
Fig.~\ref{fig:qualitative2} presents qualitative comparisons on the E-MLB dataset under daytime and nighttime conditions. EDmamba preserves structural details across diverse scenes. In the \textit{Window} and \textit{Meeting Room} examples, it retains fine architectural contours such as frame lines and corners, while suppressing background clutter. In contrast, other methods (e.g., DWF, YNoise) tend to oversmooth or break weak edges.

In nighttime \textit{Beam} scenes, EDmamba demonstrates robustness by preserving low-intensity, elongated light trails that are otherwise removed by YNoise and AEDNet. This highlights our model’s ability to distinguish faint but coherent motion patterns from stochastic noise. These results validate the model's capacity for structure-aware denoising across lighting conditions. Due to the page limit, additional qualitative results on two DND21 test scenes under two noise levels are provided in the supplementary material.
% EDmamba consistently preserves salient motion details and suppresses noise across conditions.  These qualitative results further demonstrate the robustness and generalization capability of our method across varying noise intensities and scene types.

\vspace{-0.2cm}
\subsection{Model Complexity and Efficiency Comparison}
We compare the computational efficiency of all methods in Tab.~\ref{tab:efficiency}, including FLOPs, model size, and inference time on 100K events (NVIDIA RTX A6000). Filtering-based methods (e.g., KNoise, YNoise) are fast due to their lightweight, non-learnable design but generalize poorly. In contrast, learning-based models like EDnCNN and AEDNet are slower, with larger models and inference times over 20s.

EDformer reduces model size using a lightweight Transformer backbone, but its inference is limited by the quadratic complexity of self-attention and joint modeling, taking 2.49 seconds for 100K events.  While Pre-Mamba was originally proposed for deraining, it adopts a joint 4D state-space model that entangles spatial and temporal dynamics, resulting in higher computational cost and 3$\times$ more parameters than our decoupled design.
To further validate the benefit of decoupled spatial-temporal modeling, we design a control variant named Joint-SSM as a comparative baseline.  Instead of using two separate Mamba branches, we adopt a shared Mamba block to jointly encode spatial and temporal sequences. The spatial and temporal event streams are first flattened via scan operations and concatenated before being fed into the shared Mamba. Due to the lack of task-specific specialization, this design requires more parameters and slower inference to achieve comparable performance, underscoring the necessity of noise-specific modeling.
% To further validate the benefit of this decoupling, we design a control variant named Joint-SSM as a comparative baseline, which processes both spatial and temporal features using a shared state-space block. However, this fusion implicitly entangles two physically decoupled signal patterns, leading to degraded noise discrimination in both domains. 
% To accommodate the increased feature heterogeneity, Joint-SSM requires a wider Mamba block with more channels (124.24K parameters) and slower inference (92.5ms), yet fails to improve performance (Tab.~\ref{tab:ablation}), underscoring the necessity of noise-specific modeling.
% Joint-SSM requires more parameters (124.24K) and slower inference (92.5ms), yet does not improve performance (Tab.\ref{tab:ablation}), underscoring the necessity of noise-specific modeling.
% To further explore the synergy between spatial and temporal cues, we introduce a joint variant, named Joint-SSM, where both spatial and temporal event features are processed using a shared SSM. 
% However, due to the difficulty of jointly modeling structurally distinct spatial and temporal cues, it requires larger capacity (124.24K parameters) and is slower (92.5ms), yet does not improve performance (Tab.\ref{tab:ablation}).
% However, this fusion implicitly entangles two physically decoupled signal patterns, leading to degraded noise discrimination in both domains. Joint-SSM requires more parameters and slower inference but fails to improve performance, highlighting the value of noise-specific modeling.

Our method achieves a superior trade-off between speed and capacity. With 88.98K parameters and 2.27 GFLOPs, it processes 100K events in 0.0685 seconds, which is 36$\times$ faster than EDformer and 1.4$\times$ faster than Pre-Mamba. These results highlight the effectiveness of our design in balancing efficiency and real-time applicability.
\vspace{-0.1cm}
\subsection{Ablation Experiments}
To assess the role of each component, we conduct ablation studies on the DND21 dataset (Tab.~\ref{tab:ablation}). Disabling either input feature causes performance drops, with polarity having a greater impact, as it encodes signal activity and reveals polarity-related noise unique to event data. Removing either S-SSM or T-SSM causes a larger degradation, indicating that temporal and spatial modeling plays a more critical role than feature encoding alone. Among the two branches, T-SSM contributes more in motion-heavy scenarios such as Driving, highlighting the effectiveness of temporally ordered modeling for motion continuity. We also compare against the Joint-SSM variant introduced above. Despite using more parameters and slower inference, it still underperforms our decoupled design, reinforcing that task-specific modeling is more effective than simply increasing model size.
\begin{table}[t]
% \small
\fontsize{9}{10}\selectfont
\centering
\setlength{\tabcolsep}{1.5mm}  % 压缩列间距
\begin{tabular}{lcccc}
\toprule
\multirow{2}{*}{Method Variant} & \multicolumn{2}{c}{AUC (\%)} & \multicolumn{2}{c}{$\Delta$} \\
\cmidrule(lr){2-3} \cmidrule(lr){4-5}
& Hotel-bar & Driving & Hotel-bar & Driving \\
\midrule
Full Model& \textbf{99.55} & \textbf{97.34} & -- & -- \\
 w/o Geometry  Feat. & 99.24 & 97.12 & -0.31 & -0.22 \\
 w/o Polarity  Feat. & 99.19 & 96.10 & -0.36 & -1.24 \\
w/o S-SSM & 98.78 & 95.70 & -0.77 & -1.64 \\
w/o T-SSM & 98.69 & 94.66 & -0.86 & -2.68 \\
Joint-SSM & 99.20 & 96.49 & -0.35 & -0.85 \\
\bottomrule
\end{tabular}%
% \vspace{-0.3cm}
\caption{Ablation study on DND21 at 5 Hz/pixel. We report AUC scores (\%) for two scenes (Hotel-bar and Driving), along with performance drops ($\Delta$) from the full model.}
\label{tab:ablation}
\vspace{-0.6cm}
\end{table}

\vspace{-0.3cm}
\section{Conclusion}
\label{sec:conclution}
% We present EDmamba, a lightweight, noise-aware event denoising framework that integrates spatial and temporal state-space modeling. By transforming event points into pseudo-point clouds, the Coarse Feature Extraction (CFE) module efficiently encodes both geometric and polarity-aware feature for robust event data representation. The Spatial Mamba (S-SSM) captures fine-grained spatial features and long-range dependencies, while the Temporal Mamba (T-SSM) models global temporal dynamics, ensuring effective suppression of temporally correlated noise.
% Extensive experiments demonstrate that EDmamba outperforms state-of-the-art methods in denoising accuracy, inference speed, and model efficiency, achieving real-time performance without compromising quality. Future work will focus on developing a plug-and-play version of EDmamba for deployment on resource-constrained platforms, expanding its applicability in real-world event-based vision systems.
This paper presents EDmamba, an efficient event denoising framework built on decoupled spatial and temporal state space modeling. By explicitly modeling distinct noise patterns through two specialized Mamba branches, EDmamba achieves high denoising accuracy with reduced model size and inference latency. Our Coarse Feature Extraction module captures both polarity information and geometric structure, while the decoupled design ensures efficient and task-specific processing. Extensive experiments across synthetic and real-world datasets show that EDmamba outperforms prior methods in both accuracy and efficiency. Our design highlights the value of decoupled spatiotemporal modeling for event noise, offering a new perspective on architectural specialization in event-based learning. Future work will explore integrating EDmamba into downstream systems and deployment on resource-constrained platforms, pushing forward practical and scalable event-based perception.

\bibliographystyle{ACM-Reference-Format}
\bibliography{main}
\end{document}